\documentclass[conference]{IEEEtran}

\usepackage{amsmath,amsfonts}
\usepackage{algorithm}
\usepackage{multirow}
\usepackage{tabto}
\usepackage{algpseudocode}
\usepackage{array}
\usepackage{textcomp}
\usepackage{stfloats}
\usepackage{subcaption}

\usepackage{multirow}
\usepackage{color}
\usepackage[normalem]{ulem}
\usepackage{url}
\usepackage{verbatim}
\usepackage{graphicx}
\usepackage{float}
\usepackage[font=scriptsize,justification=centering]{caption}

\def\BibTeX{{\rm B\kern-.05em{\sc i\kern-.025em b}\kern-.08em
    T\kern-.1667em\lower.7ex\hbox{E}\kern-.125emX}}
\usepackage{balance}

\DeclareUrlCommand\ULurl{%
  \renewcommand\UrlLeft{\uline\bgroup}%
  \renewcommand\UrlRight{\egroup}}
\IEEEoverridecommandlockouts
\IEEEpubid{\makebox[\columnwidth]{979-8-3503-4863-7/24/\$31.00 ~\copyright2024 IEEE \hfill}
\hspace{\columnsep}\makebox[\columnwidth]{ }}

\begin{document}
\title{Trajectory Planning of Robotic Manipulator in Dynamic Environment Exploiting DRL}
\author{\IEEEauthorblockN{Osama Ahmad}
\IEEEauthorblockA{\textit{Department of Electrical Engineering} \\
\textit{Lahore University of Management and Sciences}\\
Lahore, Pakistan \\
22060007@lums.edu.pk}
\and
\IEEEauthorblockN{Zawar Hussain}
\IEEEauthorblockA{\textit{  Department of Electrical Engineering} \\
\textit{Lahore University of Management and Sciences}\\
Lahore, Pakistan  \\
zawar.hussain@lums.edu.pk}
\and
\IEEEauthorblockN{} 
\and
\IEEEauthorblockN{Hammad Naeem}
  \IEEEauthorblockA{ \centerline{Speridian Technologies} \\
    Albuquerque, USA\\
    hammad.naeem@speridian.com}%
  }%

\maketitle
\IEEEpubidadjcol
\begin{abstract}
This study is about the implementation of a reinforcement learning algorithm in the trajectory planning of manipulators. We have a 7-DOF robotic arm to pick \& place the randomly placed block at a random target point in an unknown environment. The obstacle is randomly moving which creates a hurdle in picking the object. The objective of the robot is to avoid the obstacle and pick the block with constraints to a fixed timestamp. In this literature, we have applied a deep deterministic policy gradient (DDPG) algorithm and compared the models' efficiency with dense and sparse rewards.     
\end{abstract}


\section{Introduction}
In this modern era, robots are commonly used in many applications such as picking and placing objects, welding, surgical, agricultural sectors, and many more. Industrial robots operate in complex environments where uncertainties and causalities may happen. In some applications, humans have to physically interact with the robot which is commonly known as Human-Robot Interaction (HRI). However, this interaction can raise the risk of safety concerns for humans and the environment. 
\\
To ensure safety in industrial zones, many control strategies such as impedance control, and admittance control have been introduced. These techniques incorporate many torque or force sensors, proximity sensors, and environment sensing sensors. But these mechanisms also introduced more complexities in the robotic manipulator. Artificial potential fields are also used to avoid obstacles in the operative space of the manipulator [12], [13]. Whereas the sampling method and path planner are used to find out the trajectory of the manipulator. In these types of trajectory planning methods, the dynamics of the robot should known, or if the obstacle is moving we have to re-compute the trajectory which further adds real-time computational complexity.     
\\
With the advancement in deep neural networks (DNN) and optimization in hardware, AI techniques have been implemented in robots to perform industrial tasks in complex environments. In the last decade, many control strategies using Reinforcement Learning (RL) and neural networks have been implemented [9], [11]. 
\\ In [2] continuous control for robots using reinforcement learning has been introduced. In this literature, two major control techniques have been discussed: low-dimensionality and raw pixels using deep deterministic policy gradient (DDPG). Normalized advantage function (NAF) Q-learning algorithm shows promising results for complex systems [3].
\\
However, for convergence, RL techniques take long training time and resources. To simulate complex problems that resemble a real-time environment is still a hard problem in the reinforcement domain. In [1] the transfer learning approach to train the DRL model. The convergence will occur in fewer iterations compared to the other approaches. The learning of the manipulator can be improved by assigning efficient sparse rewards [4]. A multi-goal reinforcement environment for continuous control tasks such as pushing, sliding, and pick \& place for the 7-DOF robotic arm has been demonstrated in [5], [10], [11], [14]. Hindsight experience replay is a crucial building block that makes the training process smooth in these complex and challenging environments [4], [5]. 
\\
A hybrid model using transfer learning is applied to the Comau industrial robot, which switches between two modes [1]. For distant obstacles, it employs a Single-Query Bi-Directional Probabilistic Roadmap planner with Lazy Collision Checking (SBL). When encountering unforeseen obstacles, it uses deep reinforcement learning (DRL) to compute joint positions and velocities for evasion.
\\
This paper concentrates on trajectory planning for manipulators, aiming to navigate through an operative space while avoiding obstacles. The primary challenges lie in dealing with unknown dynamics and the movement of obstacles in any direction with unknown distributions within the operative field. The goal is to execute tasks collision-free while minimizing the magnitude of joint velocities or positions. This research explores various scenarios:  \textbf{S1)} where the target object and position randomly change with no obstacle in the operative space; and \textbf{S2)} a significant contribution in a multi-goal environment where the target object and position is randomly positioned, and the obstacle also moves randomly.
\\
The subsequent sections of this paper are structured as follows: Section II introduces methodology, Section III explains the implementation of deep reinforcement learning-based algorithms, Section IV provides simulation and results, Section V presents future study and work, and Section VI wraps up the discussions with conclusion.
\begin{figure}
    \centering{\includegraphics[width=0.48\textwidth]{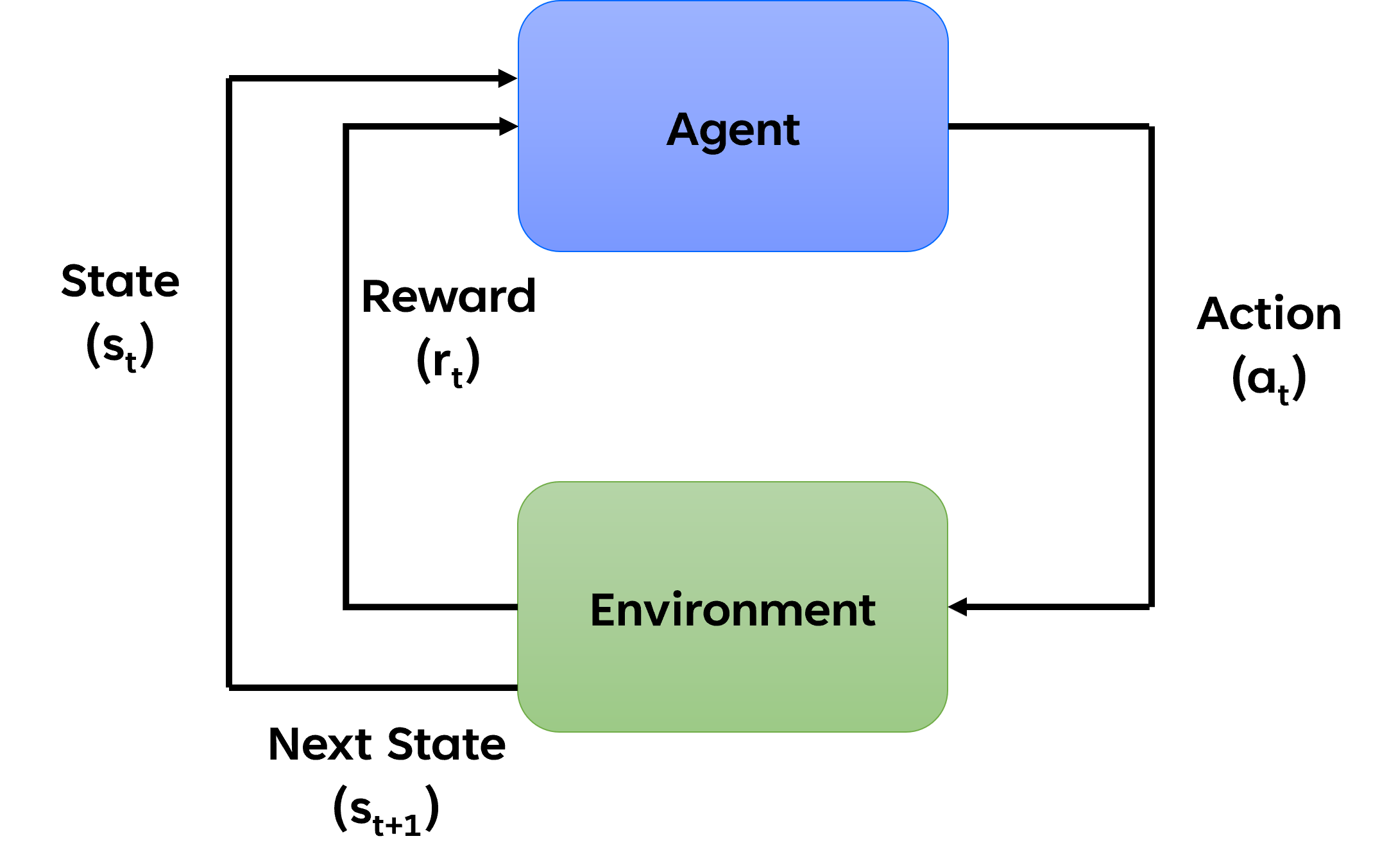}}
    \caption{Basic Diagram of Reinforcement Learning Scheme}
    \label{fig_1}
    \vspace{-0.4cm}
\end{figure}

\begin{figure*}
    \centering{\includegraphics[width=0.9\textwidth]{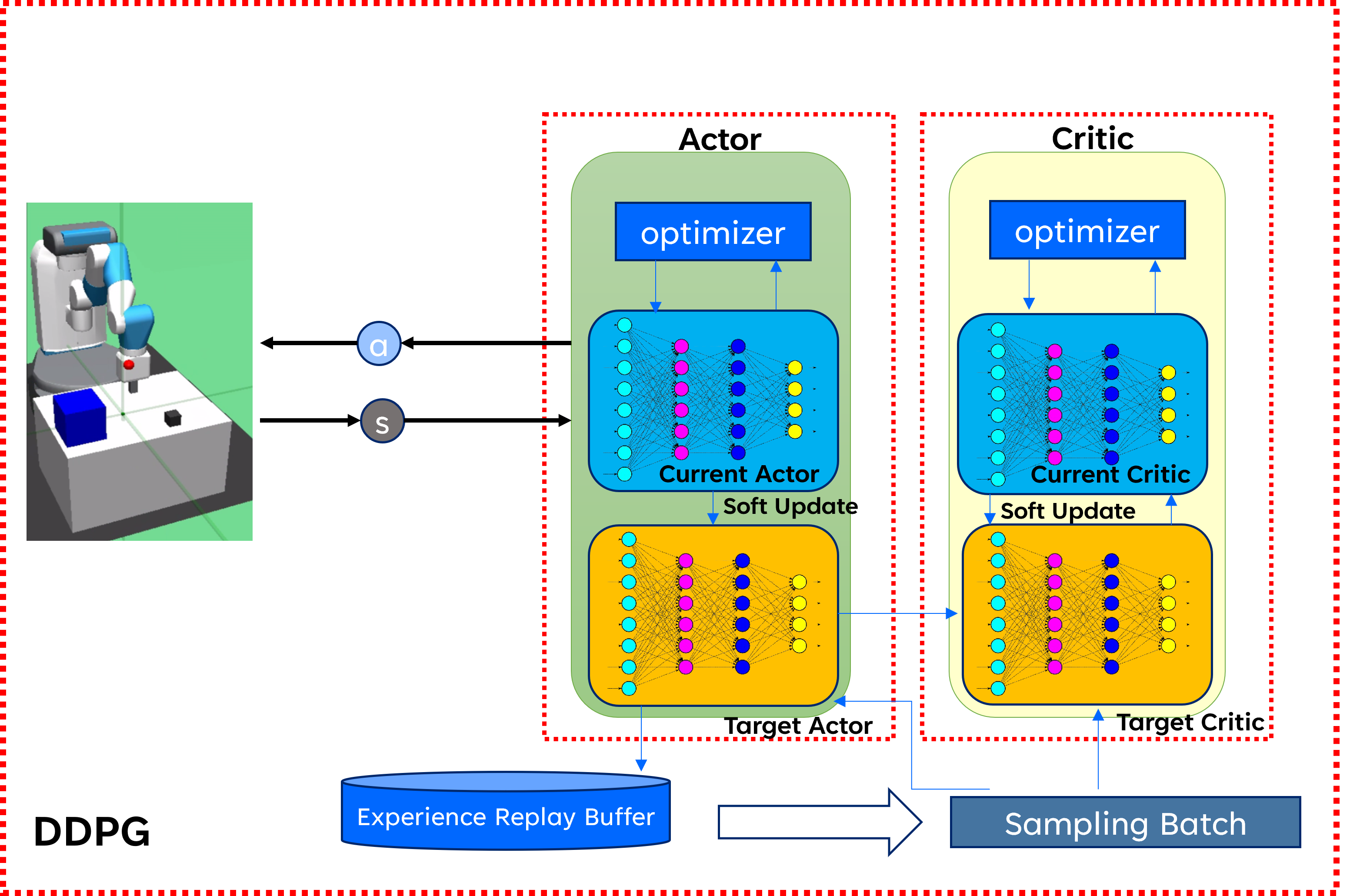}}
    \caption{Deep Deterministic Framework for Trajectory Planning for Robotic Arm }
    \label{fig_2}
\end{figure*}
\section{Methodology}
This paper presents a simulation of the robotic manipulator designed for pick-and-place tasks in an industrial setting.  The simulation considers the impact of unforeseen obstacles, such as humans or other machinery, that might disrupt the robot's planned trajectory. A camera is mounted on the manipulator that provides essential information about the surrounding environment. Moreover, the robot is equipped with sensors that track the position and velocity of the joints. 
\subsection{Modeling of Robotic Arm}
Consider a robotic manipulator having n degree of freedom with an open kinematic chain. Let joint position and joint velocity be represented by  $q,\dot{q}\in \mathbb{R}^n$. Through the position $p_e$ and orientation $\Phi_e$ of the end-effector, the joint position of the robotic arm is computed by the inverse kinematic chain. The workspace $\mathcal{W}$ is a set of all points spanned by the end-effector in a cartesian coordinates system. 
\subsection{Formulation of Control Strategies}
The formulation of sparse and dense reward is provided in eq. \ref{eq_1}  and eq. \ref{eq_2} where $R_d$ is the Euclidean distance of the end-effector from the target position, $R_o$ is the Euclidean distance of the end-effector from the obstacle. $\delta_d$ is the threshold for minimum error threshold with target position and $\delta_o$ is the maximum distance it should avoid before hitting to obstacle. $c_1$ and $c_2$ are gains for distance to target body and distance to obstacle body respectively. 
\begin{equation} 
\label{eq_1}
\boldsymbol{r_t}=-(\boldsymbol{R_d>\delta_d}+\boldsymbol{R_o < \delta_o}),
\end{equation}
\begin{equation} 
\label{eq_2}
\boldsymbol{r_t}=-(c_1\boldsymbol{R_d}+c_2\boldsymbol{R_o}),
\end{equation}

\subsection{Optimization Problem}
The trajectory planning is a minimization problem of joint velocity over the finite time horizon T as depicted in eq. \ref{eq_3}. The dynamic of a robotic arm is represented by function $f$ based on the current position and velocity. The L2-norm of joint velocity $\lVert \dot{q} \rVert$ in the objective function evaluates the minimum velocity to ensure smoothness in motion. The $c_3$ is the penalty term to control the magnitude of the joint velocity $\dot{q}$. $h(q,\dot{q})$ represents the holonomic constraints for the robotics arm. It means that the position, orientation, and as well as joint velocities will be applied within certain bounds. Through reward assignment $r_t$, it should learn how to avoid the collision with the obstacle $\mathcal{O}$ such that $\mathcal{W}(q) \cap \mathcal{O}\equiv\emptyset$. The formulated optimization problem is given by
\begin{equation}
\begin{split}
\label{eq_3}
\min_{\dot{q}}  \int_{0}^{T} r_t +\, c_3 \,\lVert \dot{q(t)} \rVert \ ^2 dt,\\   \quad s.t. \quad \ddot{q}=f(q,\dot{q}), \quad h(q,\dot{q})<=0 
\end{split}
\end{equation}
\section{Implementation of DRL based algorithm}
Reinforcement Learning (RL) enables an agent to optimize its actions through a process of trial and error within an interactive environment, guided by the feedback received from its actions and accumulated past experiences. Distinct from the principles of unsupervised learning, the primary objective in RL is to develop an action strategy that maximizes the agent's total cumulative reward. Fig. 1 illustrates the basic framework and the action-reward feedback cycle of a generic RL model.
\\
The application of trajectory planning for a robotic arm involves states S and action A which are continuous. Deep Deterministic Policy Gradient (DDPG) [2] and Normalized Advantage Function (NAF) [3] are commonly used techniques to compute policy $\pi$ with continuous space and continuous action applications. In this literature, the DDPG technique has been applied for trajectory planning and obstacle avoidance. The complete definition of the state space S and action A are defined in the Appendix.  
   \[
S = \{ \mathbf{p_e^0}, \mathbf{p_o^0}, \mathbf{p_r^0}, \mathbf{p_{obs}^0}, \mathbf{v_o^0}, \mathbf{\omega^0}, \mathbf{v_e}, d_r, d_l, v_{g_r},v_{g_l}, \alpha, \beta, \gamma \},
\]

\[
A = \{ d_{x_e},d_{y_e},d_{z_e},d_{g_r}, d_{g_l}\},
\]
DDPG is a model-free algorithm and it is based on the actor-critic approach. It is an off-policy RL algorithm based on the assumption that the system is deterministic such as for every state $s\in S$, there exists a unique action $a=\pi(s)$. The actor-critic is a two-time scale method in which the actor provides the action and the critic evaluates the action of the actor. The basic algorithm and equation have been picked up from [2].
\\
In deep reinforcement learning, the robotic environment autonomously interacts with the environment and learns how to accomplish a given task. While interacting with the environment, it stores its experience in the form of tuples $(s_t,a_t,r_t,s_{t+1})$. Majorly two neural networks are used to compute action (actor) and value (critic). The target actor and target critic are used for soft updates of the parameters. Fig. 2 demonstrates DDPG framework for trajectory planning for a robotic manipulator.
\\
Since DDPG is an off-policy method, the states-action samples are generated from the actor-network $\mu(s|\theta^\mu)$ for $T$ timestamp. Where $\mu$ is the action provided by actor-network, $s\in S$, and $\theta^\mu$ is the current weights of the network. Now, include the Ornstein–Uhlenbeck process $\mathcal{N}_t$ noise in these action values $a_t=\mu(s|\theta^\mu)+\mathcal{N}_t$ for exploration purposes. Store these experiences in the form of tuples and observe new states $s_{t+1}$. After collecting these samples from the environment, generate random samples from these tuples to train the networks. 
\\
The weights of the target actor and target critic networks are represented by $\theta^{\mu^{'}}$ and $\theta^{Q^{'}}$ respectively. Eq. 4 shows the cumulative value based on the reward and value of the next state action provided by the target critic network. The loss of the network is computed using mean-squared error in Eq. 5.  
\\
\begin{equation}
 y_i = r_i + \gamma Q'(s_{i+1}, \mu'(s_{i+1}|\theta^{\mu'})|\theta^{Q'}),
 \end{equation}
\begin{equation}
  L = \frac{1}{N} \sum_i(y_i - Q(s_i, a_i|\theta^Q))^2
\end{equation}
Eq. 6 shows how to update the actor network's parameter during back-propagation to maximize the rewards. It defines the gradient of the objective function $J$ with respect to an actor weights. The right-hand side term shows the derivative of the value of critic network $Q(s, a|\theta^Q)|_{s=s_i, a=\mu(s_i)}$ with respect to action $a$ times the gradient of the action  $\nabla_{\theta^\mu} \mu(s|\theta^\mu)$ of an actor at state $s_i$.
\\
 \begin{equation}
            \nabla_{\theta^\mu} J \approx \frac{1}{N} \sum_i \nabla_a Q(s, a|\theta^Q)|_{s=s_i, a=\mu(s_i)} \nabla_{\theta^\mu} \mu(s|\theta^\mu)|_{s_i}
\end{equation}
After computing the back-propagation at all replay batches, update the weights of the target critic and target actor network. The polyak $\tau$ is used to update the weights of target networks. It is used to soft update the weights of the network. Because of the sudden change in weights disturbs the model performance.
\begin{equation}
 \begin{split}
    \theta^{Q'} \leftarrow \tau\theta^Q + (1 - \tau)\theta^{Q'}
    \\ \theta^{\mu'} \leftarrow \tau\theta^\mu + (1 - \tau)\theta^{\mu'}
    \end{split}
\end{equation}
\begin{algorithm}[t]
\centering
\caption{DDPG algorithm [2]}
\begin{algorithmic}[1]
\State Initialize critic network $\theta^Q$ and actor $\theta^\mu$
\State Initialize target network $\theta^{Q'}=\theta^Q, \theta^{\mu'}=\theta^\mu$
\State Initialize replay buffer $R$
 \State Initialize a random process $N$ for action exploration
\For{episode $= 1, M$}
    \State Receive initial observation state $s_1$
    \For{t = 1, T}
        \State Apply action $a_t = \mu(s_t|\theta^\mu) + N_t$ 
        \State Append $(s_t, a_t, r_t, s_{t+1})$ in $R$ based on $a_t$ and $s_t$
    \EndFor
      \State Random sample the minibatch of $K$ from $R$
    \For{samples = 1, K}
        \State compute $y_i$ using Eq. 4
        \State Update critic by minimizing the loss: using Eq.5
        \State Update the actor policy using  Eq. 6
    \EndFor
    \State Update the target networks using Eq. 7  
\EndFor
\end{algorithmic}
\end{algorithm}
\section{Simulation }
\subsection{Experimental Setup}
The discussed algorithm has been implemented on a 7-DOF fetch robotic arm in the gymnasium robotics environment proposed by [5]. Mujoco is a physics engine used to simulate the robot's interaction in a given environment under some dynamics. Fig. 3 shows a snapshot of the environment, the challenge was to enable a robotic arm to pick up a block placed at random locations and track a randomly moving red target, all while avoiding unpredictable obstacles.  
\\
In every action, the robot is moved by a small displacement of the end-effector in the Cartesian coordinate system. The joint positions of the robotic arm are computed by the Mujoco framework internally using inverse kinematics. The block is uniformly distributed within a range of [-0.15, 0.15] m with reference to the initial end-effector (x,y) coordinates. Similarly, the target is defined as uniformly distributed with a range of [-0.15, 0.15] m in (x,y) coordinates, it can either be in mid-air or over the table The height of the target is sampled from a uniform distribution within a range of [0, 0.45] m. In the case of an obstacle, the x-axis is randomly distributed while the y-axis changes constantly with time. The timestamp to accomplish the task per episode is set to $T=100$. 
\\
For these experiments, four layers of neural network for actor-network have been used with $256$ hidden units. In the output layer, the $tanh$ function is used to provide normalized output. There are four layers for a critic network having a linear output layer. There are $256$ units used in each hidden layer with Relu as an activation function.
\begin{figure}[H]
\centering
\includegraphics[width=0.48\textwidth]{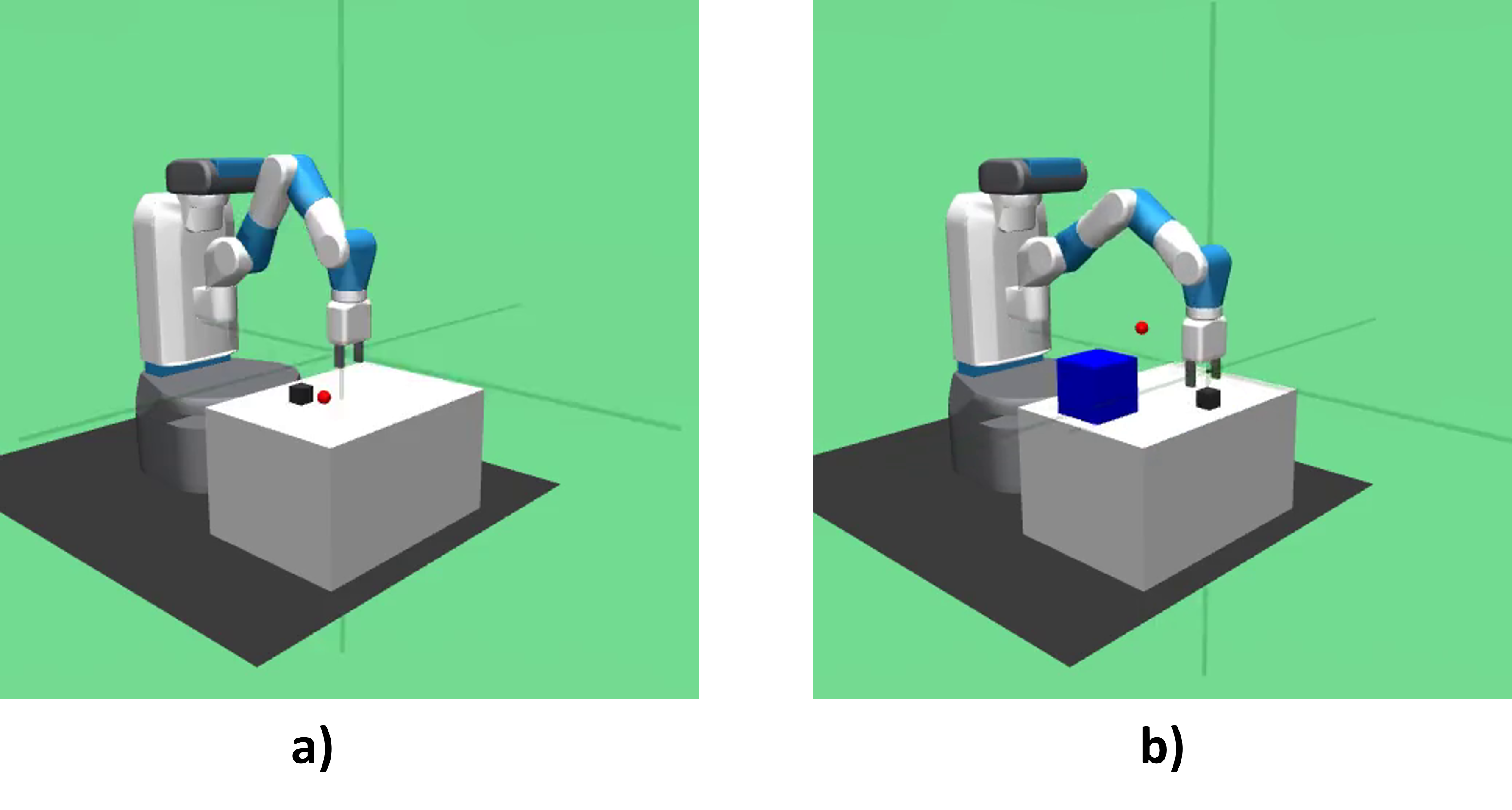}
\caption{Robotic environment in OpenAI gym a) no obstacle b) obstacle} 
\end{figure}
Experiments are conducted on a Linux machine equipped with an Intel(R) i9 $12^{th}$ Gen @ 2.40 GHz, 32 GB RAM, and NVIDIA 3080Ti GPU. The proposed method was implemented using PyTorch, specifically Mujoco and gymnasium robotics for robot operations. Each model was trained using the Adam optimizer and the hyper-parameters used for experimentation mentioned in the Appendix. 
\subsection{Results and Discussion}
We perform simulation for two case scenarios in this environment: \textbf{1)} when there is no obstacle robot has to pick the block and reach the target \textbf{2)} obstacle is randomly moving, block and target are randomly placed. In case 1, the success rate is measured that the robot has accomplished the task. In case 2, the success rate is measured by the robot having achieved its task and not colliding with an obstacle. Fig. 4 a) shows the success rate after every training episode for case 1 when a reward is sparse and Fig. 4 b) shows the success rate when a reward is dense. It is depicted that the model with sparse reward converges in a few episodes compared with dense reward.
\begin{figure}[H]
\centering
\includegraphics[width=0.52\textwidth]{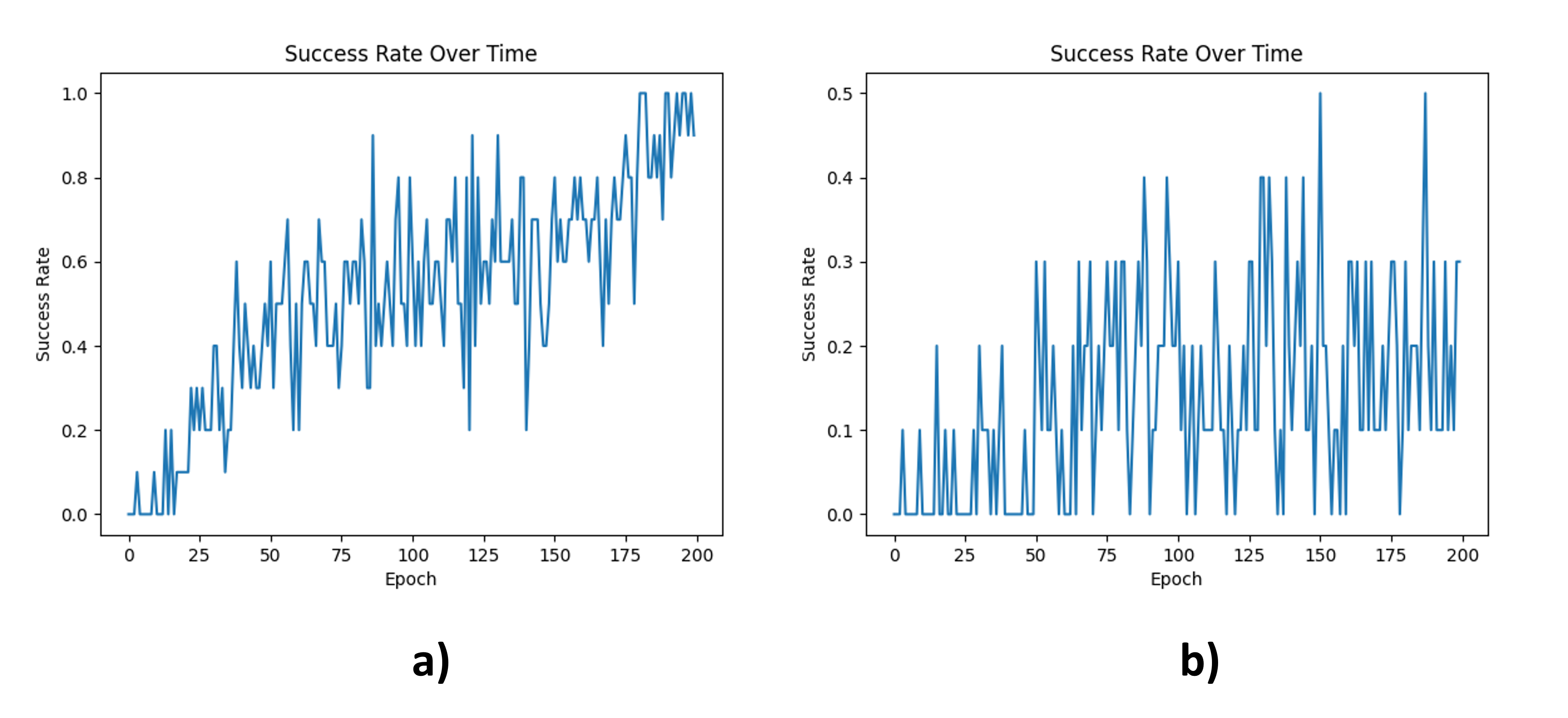}
\caption{Success rate with no obstacle a) sparse reward b) dense reward}
\end{figure}
Fig. 5 a) and b) represent the training response of the actor-network for sparse and dense rewards respectively. In the case of sparse reward, agents receive only binary values which results in more exploration hence actor loss increases sharply at the beginning. In contrast, the dense rewards agent exploits its knowledge of the environment more consistently because of the difference in distance of the agent's actions, allowing it to refine its policy with less exploration. Dense reward results in more quicker and stable learning with lower loss values, suggesting a more efficient learning process than the sparse reward case.
\begin{figure}[h]
   
\includegraphics[width=0.52\textwidth]{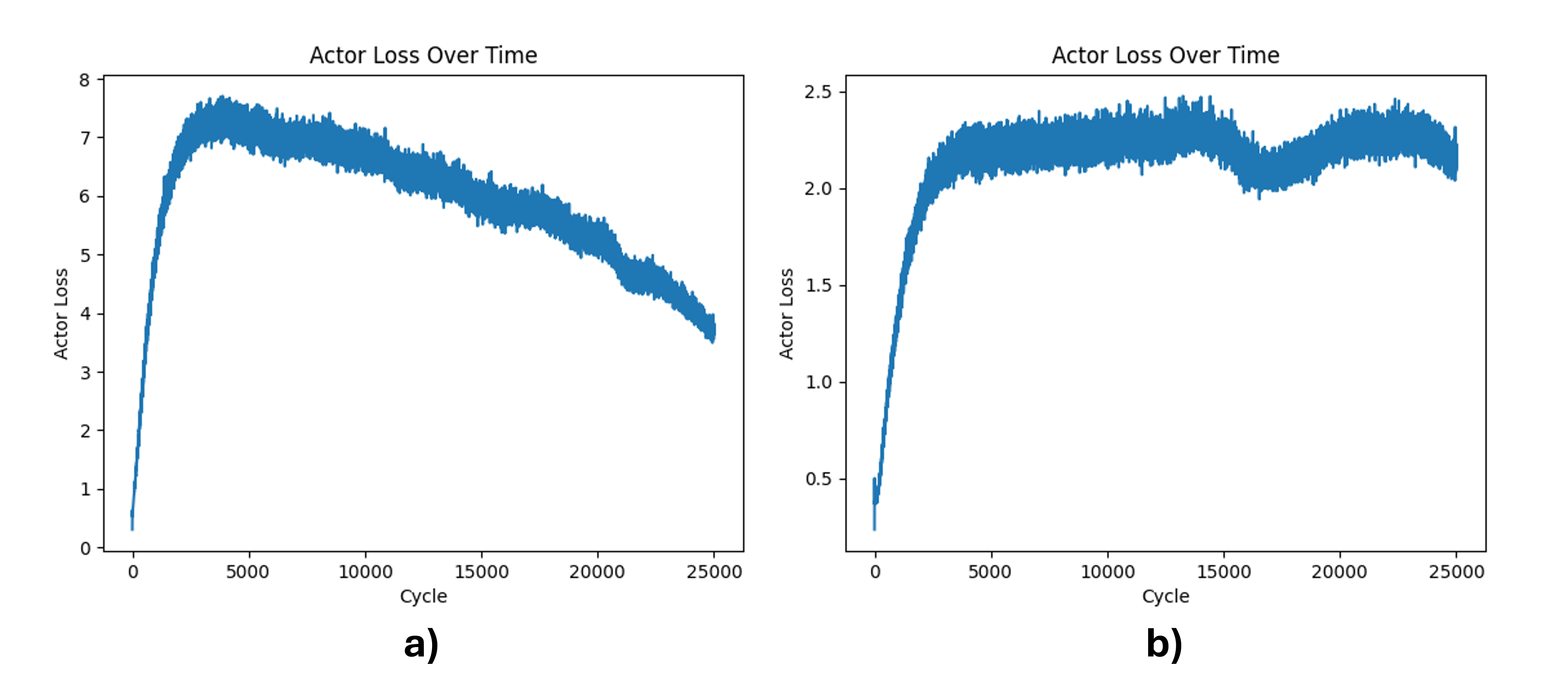}
\caption{Actor loss when an obstacle is moving with a) sparse reward b) dense reward}
\end{figure}

The training curve of critic loss with sparse and dense rewards is shown in Fig. 6. The higher loss in the case of sparse reward leads to exhibits the difficulty for the critic in predicting value as opposed to the critic in dense reward where the agents appear to learn more efficiently indicated by the steeper initial drop and overall lower loss values.  In general, the critic's task is more complex as compared to the actor's because it evaluates the actions based on rewards which can vary widely as the actor explores the environment.   
\begin{figure}[h]
\includegraphics[width=0.52\textwidth]{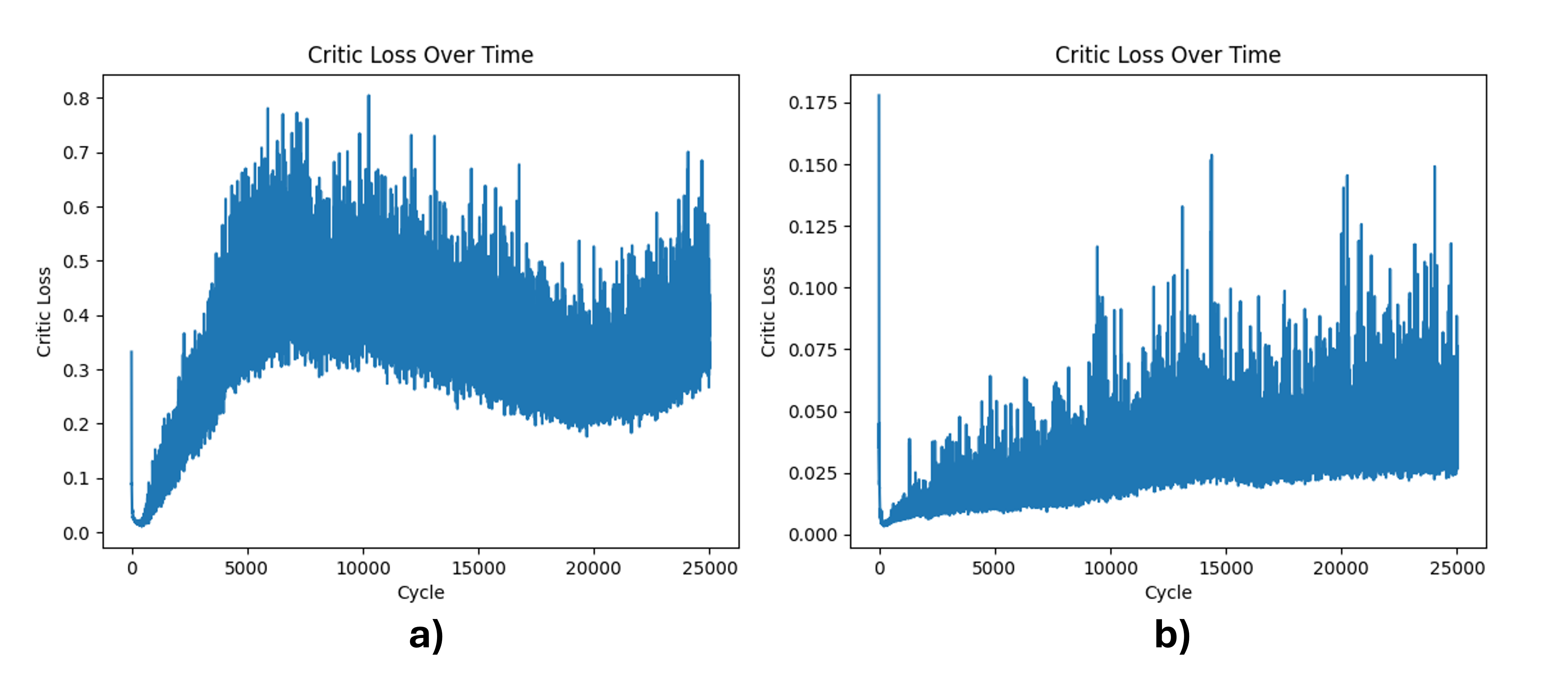}
\caption{Critic loss with an obstacle when a) sparse reward b) dense reward }
\end{figure}

The performance of the model has been evaluated on the test set, it consists of 40 random cases. The success rate with sparse and dense rewards is shown in Fig. 7, the results indicate the average test results of random cases after each training epoch.  The response of success rate shows that it is increasing after every episode. The model with sparse reward shows overall good performance in comparison to the dense reward. The model's efficiency is determined by computing the success rate on $10^5$ random test cases. Table I shows the results of the training on different cases. Case 1 with dense reward shows poor performance when a model is trained on $200$ iterations. If this model is trained further, it can show significant performance improvement. But in this analysis, we are comparing reward functions for fixed episodes.
\begin{figure}[H]
\includegraphics[width=0.5\textwidth]{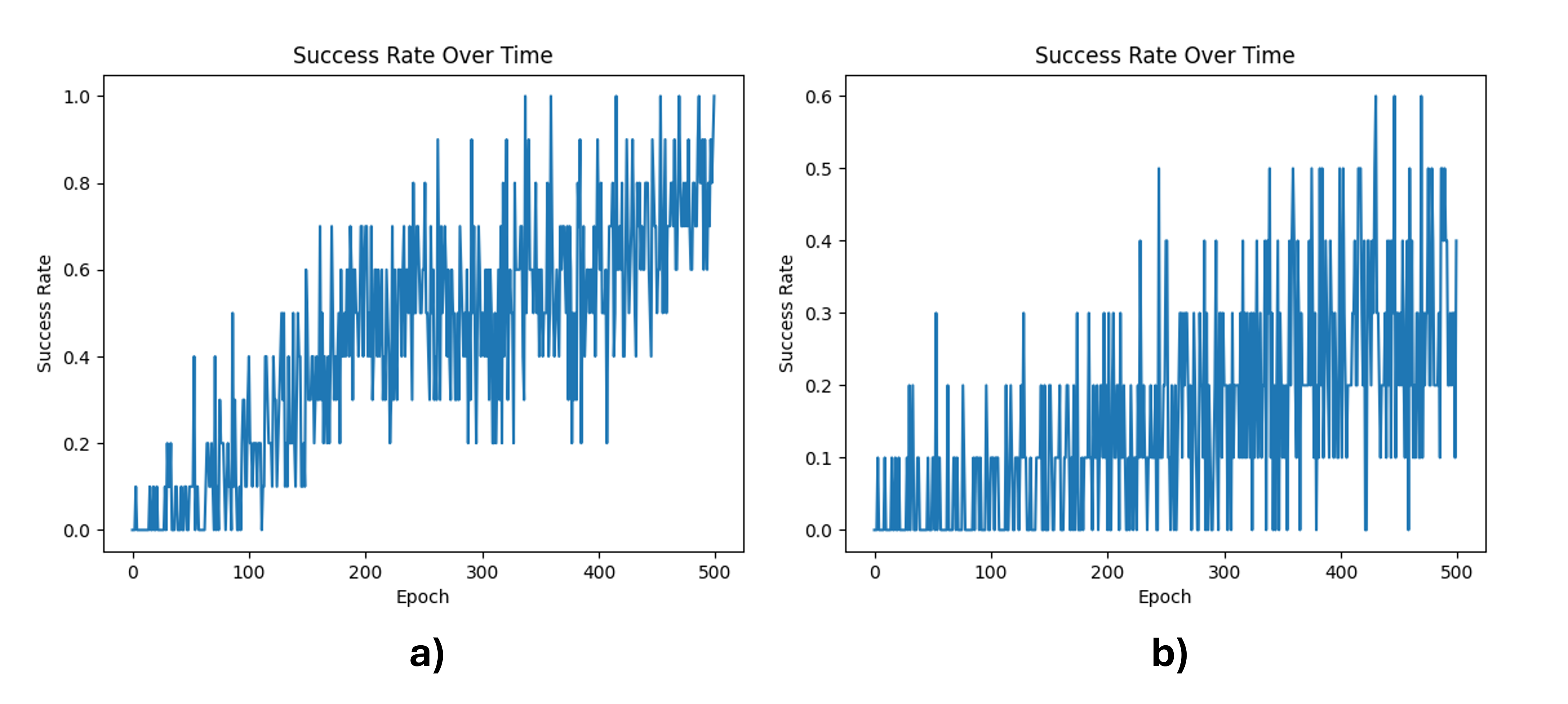}
\caption{Success rate with an obstacle when a) sparse reward b) dense reward}
\end{figure}
 It has been observed that the failure cases happen because of two main reasons: the number of steps is limited and if the block falls below the table during pick, the block falls beyond the workspace of the robot. When the reward is sparse for case 2, DDPG shows promising results. With the dense reward, the model didn't converge even for $500$ training epoch. Because the task's complexity is relatively increased. The sparse reward introduces flexibility in accomplishing the task. The collision with obstacles can be improved further. 
\begin{table}[H]
\centering
\caption{Evaluation on test cases}
\label{Table_2}
\begin{tabular}{ | m{8em} | m{5em}| m{6em} | m{6em}|} 
 
  \hline
   \textbf{Cases}&\textbf{Success Rate(\%)} & \textbf{Fail to reach the target position (\%)} & \textbf{Collision with obstacle (\%)} \\
   \hline
   Case 1 (sparse reward) & $94.265$ & $5.735$ & $NA$\\
   \hline
   Case 1 (dense reward) & $36.03$ & $63.97$ & $NA$\\
   \hline
    Case 2 (sparse reward) & $71.844$ & $9.618$ & $18.538$\\
   \hline
    Case 2 (dense reward) & $16.477$ & $83.271$ & $0.252$\\
   \hline
\end{tabular}
\end{table}
\section{Future Study and Work}
Since DDPG uses linear layers in a deep neural network, it works with a fixed number of observations. But in the current scenario, due to the random nature of the obstacle, it can appear at any instance in the operative field of the robot. This problem can be visualized as a dynamic graph where the obstacle (node) appears at any time, so the robot's behavior will change. Similar approaches like  [6], [7], and [15] can be applied to these problem statements. These methods used Graph Neural Networks (GNN) and Reinforcement Learning (RL) to explore the environment or accomplish goals. Another approach is also commonly used to combine Model Predictive Control (MPC) and Reinforcement Learning (RL) in path planning applications [8]. It solves the mixed static and dynamic obstacle avoidance problem in an unknown environment.
\section{Conclusion}
This work successfully implements a reinforcement learning-based trajectory-finding algorithm in a complex and unknown environment. It has been found that sparse rewards make learning smooth and the performance efficiency is better than with a model having dense rewards. In the moving obstacles environment, a deep deterministic policy gradient shows a good response. The robot prioritized obstacle avoidance over picking the block in finite steps.

\appendix
\subsection{Notation Definition}
The notations in state space S are defined as: 
\\
$\mathbf{p_e^0}:$ position of end-effector w.r.t global frame\\
$\mathbf{p_o^0}:$ position of object w.r.t global frame\\
$ \mathbf{p_o^g}:$ Position of block w.r.t gripper\\
$ \mathbf{p_{obs}^0}:$ position of obstacle w.r.t global frame\\
$\mathbf{v_o^0}:$ velocity of obstacle w.r.t global frame\\
$\mathbf{\omega^0}:$ angular velocity of block w.r.t global frame\\
$ \mathbf{v_e}:$ velocity of end-effector\\
$d_r:$ displacement of right side gripper\\
$d_l:$ displacement of left side gripper\\
$v_{g_r}:$ velocity of right side gripper\\
$v_{g_l}:$ velocity of left side gripper\\
$\alpha:$ global x rotation of a block in XYZ Euler frame rotation\\
$\beta:$ global y rotation of a block in XYZ Euler frame rotation\\
$\gamma:$ global z rotation of a block in XYZ Euler frame rotation\\
\\
The notations in action space A are defined as: 
\\
$d_{x_e}:$ displacement of end-effector along x axis\\
$d_{y_e}:$ displacement of end-effector along y axis\\
$d_{z_e}:$ displacement of end-effector along z axis\\
$d_{g_r}:$ displacement of right side gripper\\
$d_{g_l}:$ displacement of left side gripper\\

\subsection{HyperParameter}
The list of parameters when an obstacle is moving are:
\\
critic learning rate: $0.0001$ \\
Actor learning rate: $0.0001$ \\
buffer size: $int(1E6)$ \\
polyak: $0.96$ \\
L2 action penalty: $1.0$\\
No. of observation clip: $200.$ \\
gamma: $0.98$ \\
epochs: $500$ \\
cycles per epoch: $50$  \\
rollout batch size per mpi thread: $2$  \\
training batches per cycle: $40$ \\
batch size: $256$ \\ 
  number of test rollouts per epoch: $10$ \\
percentage of time a random action: $0.3$\\
noise: $0.2$\\
replay strategy: `future',  supported modes: future, none\\
replay k: $4$,  number of additional goals used for replay, only used if off policy data=future\\
    
\end{document}